\documentclass[10pt,twocolumn,letterpaper]{article}

\usepackage{wacv} 

\usepackage[utf8]{inputenc} 
\usepackage[T1]{fontenc}    
\usepackage{url}            
\usepackage{booktabs}       
\usepackage{amsfonts}       
\usepackage{float}          
\usepackage{nicefrac}       
\usepackage{microtype}      
\usepackage{xcolor}         
\usepackage{graphicx} 
\usepackage{makecell}

\definecolor{wacvblue}{rgb}{0.21,0.49,0.74}
\usepackage[pagebackref,breaklinks,colorlinks,allcolors=wacvblue]{hyperref}
\definecolor{dgreen}{rgb}{0.0, 0.5, 0.0}

\def\wacvPaperID{1101} 
\def\confName{WACV}
\def\confYear{2026}

\title{Direct Visual Grounding by Directing Attention of Visual Tokens}

\author{Parsa Esmaeilkhani\\
Temple University\\
Philadelphia, USA\\
{\tt\small parsa.esmaeilkhani@temple.edu}
\and
Longin Jan Latecki\\
Temple University\\
Philadelphia, USA\\
{\tt\small latecki@temple.edu}
}

\graphicspath{ {./figures/} }

\begin{document}
\maketitle

\begin{abstract}
Vision Language Models (VLMs) mix visual tokens and text tokens.
A puzzling issue is the fact that visual tokens most related to the query receive little to no attention in the final layers of the LLM module of VLMs from the answer tokens,
where all tokens are treated equally, in particular, visual and language tokens in the LLM attention layers.
This fact may result in wrong answers to visual questions, as our experimental results confirm.
It appears that the standard next-token prediction (NTP) loss provides an insufficient signal for directing attention to visual tokens.
We hypothesize that a more direct supervision of the attention of visual tokens to corresponding language tokens in the LLM module of VLMs will lead to improved performance on visual tasks.
To demonstrate that this is indeed the case,
we propose a novel loss function that directly supervises the attention of visual tokens.
It directly grounds the answer language tokens in images by directing their attention to the relevant visual tokens.
This is achieved by aligning the attention distribution of visual tokens to ground truth attention maps with KL divergence.
The ground truth attention maps are obtained from task geometry in synthetic cases or from standard grounding annotations (e.g., bounding boxes or point annotations) in real images, and are used inside the LLM for attention supervision without requiring new labels.
The obtained KL attention loss (KLAL) when combined with NTP
encourages VLMs to attend to relevant visual tokens while generating answer tokens.
This results in notable improvements across geometric tasks, pointing, and referring expression comprehension on both synthetic and real-world data, as demonstrated by our experiments.
We also introduce a new dataset to evaluate the line tracing abilities of VLMs. Surprisingly, even commercial VLMs do not perform well on this task. Data and code: \href{https://github.com/ParsaKhani/visual-grounding-datasets}{GitHub}.
\end{abstract}

\vspace{-5mm}
\section{Introduction}
\vspace{-2mm}

Vision-Language Models (VLMs) have achieved remarkable success in various multimodal tasks, including image captioning, visual question answering (VQA), and image-text retrieval. Models like CLIP~\cite{radford2021learning}, Flamingo~\cite{alayrac2022flamingo}, Llava \cite{liu2023visual}, MiniGPT4 \cite{zhu2023minigpt}, and Qwen-VL \cite{bai2023qwenvlversatilevisionlanguagemodel} have demonstrated the efficacy of aligning visual and textual modalities through contrastive and generative pretraining strategies. However, despite their impressive performance on general benchmarks, VLMs often struggle with tasks requiring intricate visual reasoning, such as spatial relations~\cite{kamath2023s, VisualGroundingCVPR2024}, object counting~\cite{johnson2017clevr,qharabagh2024lvlm}, and visual inference~\cite{hudson2019gqa, fu2025hidden, tong2024eyes}.
But more alarming is the fact that 
VLMs struggle with simple, low-level vision tasks 
like whether two lines intersect or two geometric primitives overlap or are close together \cite{rahmanzadehgervi2024vision}.
This seems counterintuitive since VLMs excel in complex visual tasks, like a detailed description of image content \cite{cheng2025caparena, Rotstein_2024_WACV, garg2024imageinwords}.

\begin{figure}[htbp]
  \centering
  \includegraphics[width=0.9\linewidth]{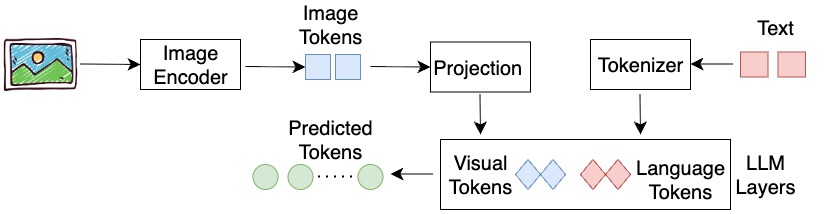}
  \caption{Processing flow in VLMs}
  \label{fig:improc}
\end{figure}

Fig.~\ref{fig:improc} illustrates the processing flow of recent VLMs,
like LLava-v1.5 \cite{liu2023visual} and Qwen2.5-VL \cite{bai2025qwen2}.
The input image (or images) are first passed through a frozen visual encoder (typically a ViT, CLIP, or DINOv2) and then mapped with a projection layer (alignment module) to a language token embedding space.
Finally, the image tokens are passed to an LLM,
where they are mixed with special and language tokens. 
The mixing is performed in the attention layers of the VLM.
We call the LLM tokens corresponding to the input image tokens visual tokens. 

As visual tokens are transformed through the layers of an LLM, their embeddings change.
The goal is to better align them with the embeddings of the language tokens to yield the desired answers.
However, there is a danger that
the information from visual tokens gets lost among all LLM's tokens after they are processed by the LLM layers.
In an extreme case, the visual tokens are ignored and the LLM can even hallucinate an image description for an empty image, as was demonstrated in  \cite{payattentionimageECCV2024,tong2024cambrian}.
Further evidence that visual tokens are often ignored among all LLM tokens is the fact that after removing half of the visual tokens, the VLM performance does not decrease \cite{chen2024imageworth12tokens}. 
As pointed out in \cite{fu2025hidden},
\emph{The LLM’s ability to use its vision representations is a limiting factor in VLM performance.}

\textbf{So, the problem is the attention mixing of visual and language tokens in the LLM module of VLMs,
where all tokens are treated equally, in particular, visual and language tokens.} 
Indeed, answer tokens in state-of-the-art VLMs allocate only a small fraction of their attention to visual tokens (see Section 3.1 of the supplementary material).
We call this problem the \textbf{problem of attention to visual tokens}.
Intuitively, we would expect high attention of tokens representing language concepts to the corresponding regions in images,
e.g., the language token "cat" should pay high attention to the tokens representing the image region of the cat. 
As our experimental results demonstrate,
the standard next-token prediction (NTP) loss provides only a weak and indirect signal for directing attention to visual tokens.
The issue is that LLMs have difficulty recognizing the special role of visual tokens in answering image-related questions and often rely on language priors instead \cite{wang2024mllm}.

The issue persists even if visual grounding is utilized.
Visual grounding involves localizing a specific object (or a group of objects) in an image referred to with a natural language expression.
This can be done with a bounding box containing the object or with an object mask pointing to the object location, e.g., Pix2Seq \cite{pix2seqICLR2022} and Kosmos-2 \cite{kosmos2-ICLR2023}.
Many visual grounding approaches are able to accurately locate objects, e.g., with bounding boxes, but the question is whether they really know the location of these objects in images, i.e., do they know which visual tokens represent the corresponding object regions (ROIs) in the image. 
For example, 
the attention visualization experiments in \cite{VisualGroundingCVPR2024} demonstrate that it is often not the case.


\begin{figure}[H]
    \centering
    \begin{subfigure}[b]{0.23\textwidth}
        \centering
        \includegraphics[width=\textwidth]{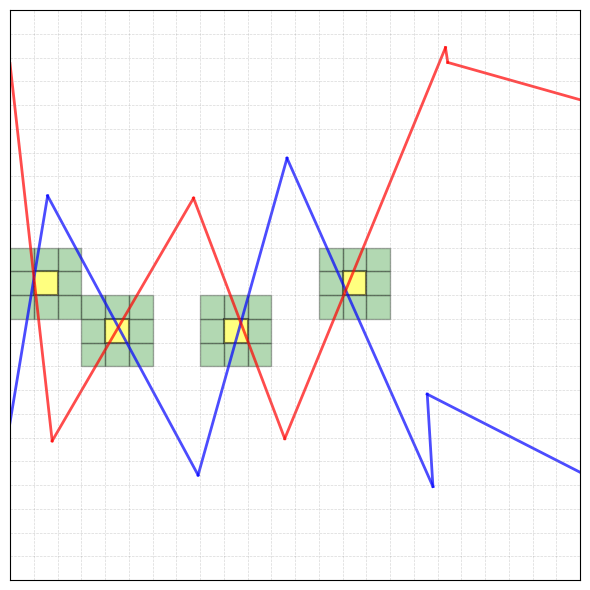}
        \caption{Line Intersection task}
        \label{fig:subfig_patch}
    \end{subfigure}
    \hfill
    \begin{subfigure}[b]{0.23\textwidth}
        \centering
        \includegraphics[width=\textwidth]{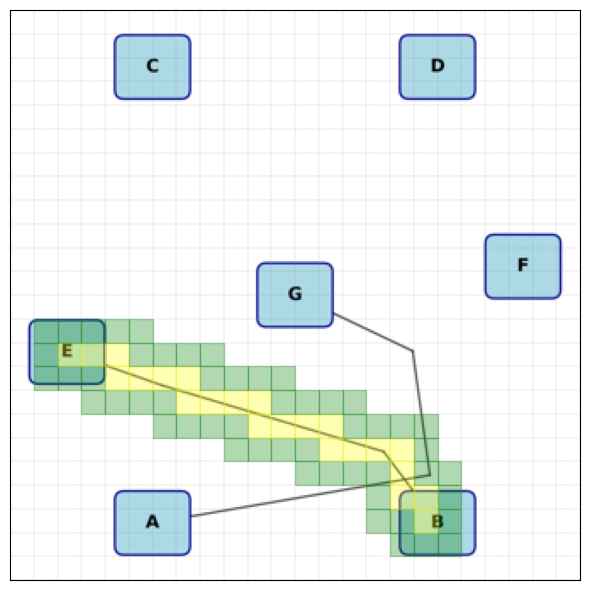}
        \caption{Is node E connected to node B?}
        \label{fig:subfig_graph}
    \end{subfigure}
    \caption{Visualization of the ground truth map for the attention of answer tokens. Yellow highlights the target patches with the highest mass of the distribution, while green indicates the surrounding neighbors. The other patches have a fixed low value.}
    \label{fig:intersections_combined}
\end{figure}

\textbf{
Our main hypothesis is that direct supervision of the attention of language tokens to corresponding visual tokens in the LLM module of VLMs will lead to improved performance on visual tasks.}
To demonstrate that this is indeed the case,
we propose a novel training framework that directly supervises the attention maps of VLMs using a combination of Kullback-Leibler (KL) divergence loss in addition to the next-token prediction loss. 
The proposed KL attention loss (KLAL) directly grounds the answer text tokens in relevant visual tokens by increasing the attention of the answer tokens to the relevant visual tokens.
By aligning the model's attention distributions with ground truth (GT) maps with KL divergence, the proposed KLAL strengthens the direct links between language and corresponding visual tokens.
In our framework, the ground-truth (GT) attention maps could come from task geometry in the case of synthetic datasets, and from standard grounding annotations (e.g., bounding boxes or point annotations) in the case of real images, projected onto image patches with a smoothing function. This way, no new labels are required, while the maps provide explicit supervision at the token level. Our contribution is not in collecting new annotations, but in introducing a simple way to incorporate these GT attention maps into the LLM’s training.
Fig.~\ref{fig:intersections_combined} illustrates GT target attention maps for two tasks. For line intersection, the attention highlights patches with intersection points, while for line tracing, it follows the patches along the path connecting the queried nodes. In both cases, KLAL directs the answer tokens to focus on the relevant visual patches, improving both attention and answer quality as shown in Fig.~\ref{fig:multi_task_attention}.


Our approach is model-agnostic and can be seamlessly integrated into existing VLMs without architectural modifications.
It is also simple to implement in that it does not require attaching any task-specific additional heads (e.g., for object localization or segmentation).
As our experimental results demonstrate, the explicit visual grounding with KLAL not only improves the quality of VLM answers but also 
improves the deep embeddings of visual tokens.

Attention visualization has also been used to explain transformer inference,  e.g., in \cite{darcet2023vision} for a vision transformer (ViT). 
As argued in \cite{LLM-not-consistent2025},
it helps to ensure that LLMs provide correct and consistent information.
So, the proposed KLAL can also be helpful in improving the interpretability of VLM responses.
Our main contributions are as follows:
\begin{itemize}
    \item Introduce an auxiliary loss to direct the attention of language answer tokens to visual tokens representing the relevant parts of the image.
    \item Provide clear experimental evidence that the improvement in the attention to relevant parts of the image contributes to the performance increase on visual tasks. This yields better explainability of the results, which is visible in the presented attention maps. 
    \item  Demonstrate that the deep embeddings of visual tokens improve, as tokens with high-norm concentrate in the parts of the image most relevant to the correct answers.
    \item Introduce a new dataset to evaluate and finetune the performance of VLMs on a Line Tracing task that is essential for knowing which objects are connected in a given image. 
    \item Evaluate the performance of open source and commercial VLMs on a variety of geometric and visual grounding tasks in order to determine their fundamental visual abilities. The specific tasks are counting the number of intersection points of lines, identifying which objects are connected by lines, pointing to object locations, and resolving referring expressions, where the model must map a natural language description to the correct object in the image.
\end{itemize}

\vspace{-3mm}
\section{Preliminaries}
\vspace{-2mm}

LLMs generate tokens in an autoregressive fashion using a decoder-only Transformer with causal masking \cite{radford2018improving}. At each step \(t\), the model attends only to tokens at positions \(<t\) and selects the most probable next token from its vocabulary. Then, it updates its weights by minimizing the next-token prediction loss. Our visual attention loss directly supervises the attention weights on visual tokens with respect to other query tokens.

\subsection{Language Modeling}
During training of VLMs, the entire sequence of input tokens typically follows the structure: text tokens corresponding to the system prompt (\(\mathbf{X}_{\mathrm{sys}}\)), followed by visual tokens extracted from the image (\(\mathbf{X}_V\)), and finally text tokens representing the instruction or question provided after the image (\(\mathbf{X}_{\mathrm{instruct}}\)).
These segments are concatenated to form the full input context:$$
\mathbf{X}
=
\bigl[\,
\mathbf{X}_{\mathrm{sys}},\;\mathbf{X}_V,\;\mathbf{X}_{\mathrm{instruct}}
\bigr]
$$
The image is processed through the pre-trained vision encoder and mapped to visual embedding tokens which are then mapped to language embedding space and mixed with language tokens. All of these tokens are passed through the LLM backbone, which then generates an answer sequence \(\mathbf{X}_{a}=(x_{1},\dots,x_{T_a})\) token-by-token under a left-to-right causal mask.  The standard next-token prediction loss is:
\begin{equation}
L_{\mathrm{NTP}}(\theta)
= -\frac{1}{T_a}\sum_{i=1}^{T_a}
\log p_{\theta}\bigl(x_{i}\mid \mathbf{X},\,\mathbf{X}_{a,<i}\bigr)
\end{equation}
where \(\theta\) denotes the model parameters and \(T_a\) is the length of the answer sequence.  At each step \(i\) the model predicts \(x_i\) conditioned on the
full context \(\mathbf{X}\) and the previously generated answer tokens
\(\mathbf{X}_{a,<i}\).

\subsection{Attention Block}
The attention matrix of a single attention head $h$ at layer $l$ is given by:
\begin{equation}
\mathbf{A}^{(l, h)} = \mathrm{softmax} \left( \frac{\mathbf{Q}_h \mathbf{K}_h^\top}{\sqrt{d_k}} \right)
\end{equation}
where $\mathbf{Q}_h, \mathbf{K}_h \in \mathbb{R}^{n \times d_k}$ are the query and key matrices, and $d_k$ is the head dimension. The softmax normalizes each row of the matrix such that the attention scores sum to 1. 
As each Transformer layer has \(H\) attention heads in parallel, the head outputs are concatenated along the
feature dimension to form the layer’s multi-head representation.
The attention matrices from each layer are reused in our
auxiliary visual attention loss which aims at encouraging the model to focus on
semantically relevant image regions.

\vspace{-3mm}
\section{Methodology}
\vspace{-2mm}

The main idea of the proposed approach is to extend the training (text, image) pairs by adding target ground truth (GT) attention maps and utilizing a loss function to compare the attention of visual tokens (with respect to the text answer tokens) to the GT attention maps.
We treat both attentions as distributions and compare them with KL divergence. So, we call our loss function KLAL.
The GT attention maps are constructed automatically, i.e., no manual labeling is necessary.
When used in addition to NTP, KLAL helps the model focus more on regions in the image that are decisive for the given task. It does so by increasing the attention on visual tokens corresponding to regions of interest during finetuning.

A first step is to compute the attention distribution over the visual tokens with respect to answer tokens.
Let \(\mathcal{S}=(\mathbf{X},\mathbf{X}_a)\) denote one training sample containing the full input context (system + visual + instruction tokens) and the generated answer token sequence $\mathbf{X}_a$.
The attention matrices \(\mathbf{A}_h\) can be used to compute the attention distribution of the visual tokens to the specific answer tokens. 
We use the last generated answer, as it captures a summary of the model’s focus and reflects how information has been aggregated across the preceding tokens.
Let \(\alpha^{(l,h)}\) be the submatrix of $\mathbf{A}^{(l ,h)}$ representing the attention of all visual tokens to the last answer token for layer \(l\) and head \(h\). 
This submatrix is a slice of the last row of the attention matrix. We normalize \(\alpha^{(l,h)}\) to sum to one so that it represents a probability distribution. 
We average the normalized submatrices \(\alpha^{(l,h)}\) across heads to obtain a single distribution:
\begin{equation}
Q^{(l)}_i(\mathcal{S})
= \frac{1}{H}\sum_{h=1}^H \alpha^{(l,h)}_{i}(\mathcal{S}),
\quad i \in I_V,
\end{equation}
where \(I_V\) denotes the set of indices corresponding to the visual tokens and $l$ is the index of the LLM layers.

Let \(P(\mathcal{S})\) be a GT target distribution over visual tokens $I_V$, which is defined at the end of this section. 
Its goal is to increase the model’s focus on semantically important visual patches, i.e., the patches that are most relevant for obtaining correct answers.

We introduce a novel attention loss based on KL divergence (KLAL)
to bring the predicted distribution \(Q^{(l)}(\mathcal{S})\) at each layer \(l\) closer to constructed GT distribution \(\mathcal{P(S)}\). 
\begin{equation}
\begin{aligned}
\mathcal{L}_{\mathrm{KLAL}} &= \frac{1}{L} \sum_{l=1}^L D_{\mathrm{KL}}\bigl(P(\mathcal{S}) \,\|\, Q^{(l)}(\mathcal{S})\bigr) \\
&= \frac{1}{L}\sum_{l=1}^L \sum_{i \in I_V} P_i(\mathcal{S}) \log\left(\frac{P_i(\mathcal{S})}{Q^{(l)}_i(\mathcal{S})}\right)
\end{aligned}
\end{equation}
where both distributions are defined over the set of visual tokens $I_V$ and
L denotes the total number of layers in LLM.

Finally, we combine the next-token prediction loss with our visual attention loss to optimize the VLM's parameters for both objectives:
$\mathcal{L}_{\text{total}} = \mathcal{L}_{\text{NTP}} + \lambda \mathcal{L}_{\text{KLAL}}.$
We empirically set $\lambda$ to 1, which we found effective across all tasks (see Sec. 4 in supplementary material for ablations on $\lambda$ and head/layer design choices).
The target GT distribution $P(\mathcal{S})$ provides guidelines for LLM regarding which vision patches to focus on when answering the question in $\mathcal{S}$. 
$P(S): I_V \to [0, 1]$ is defined as
\begin{equation}
P(\mathcal{S}) = \mathrm{Normalize}\Bigl(\mathrm{Smooth}\bigl(\mathbf{1}(I_P)\bigr)\Bigr),
\end{equation}
where \(P(\mathcal{S})\) is normalized to sum to one; \(\mathrm{Smooth}(\cdot)\) is any smoothing function (e.g., Gaussian); \(\mathbf{1}(I_P)\) is an indicator equal to 1 for target patches \(I_P\subseteq I_V\) and 0 otherwise.


The definition of the set of target patches \(I_P\) that induces GT maps is task-specific. For example, in the task of counting the number of intersection points between two polygonal curves, the patches containing the intersection points constitute the target patches. These are marked in yellow in Fig.~\ref{fig:intersections_combined}(a), where the green patches are obtained by smoothing. 
In Fig.~\ref{fig:intersections_combined}(b),
the yellow patches trace the line connecting nodes E and D.
They illustrate the target patches \(I_P\) for the answer "Yes" to the query
"Is node E connected to node D?". 

For real images, the GT map construction is based on existing annotations. For point annotations at object centers, \(I_P\) is the patch containing the point with light smoothing around it. For bounding-box annotations describing the referred object, we take the box’s center line, vertical or horizontal depending on the box orientation, and mark the patches it traverses. Although \(I_P\) varies by task, our pipeline builds it automatically without manual labeling (Sec. 1.5 in supplementary material). For more complex visual tasks, without explicit annotations, GT maps can come from weakly supervised grounding methods \cite{strudel2022weaklysupervisedsegmentationreferringexpressions, liu2019adaptive, shaharabany2023similarity}, providing pseudo-labels at scale for our KLAL in real-world applications.

\begin{figure*}[!htbp]
\setcounter{subfigure}{0}
\begin{subfigure}[b]{0.23\textwidth}
  \includegraphics[width=\textwidth]{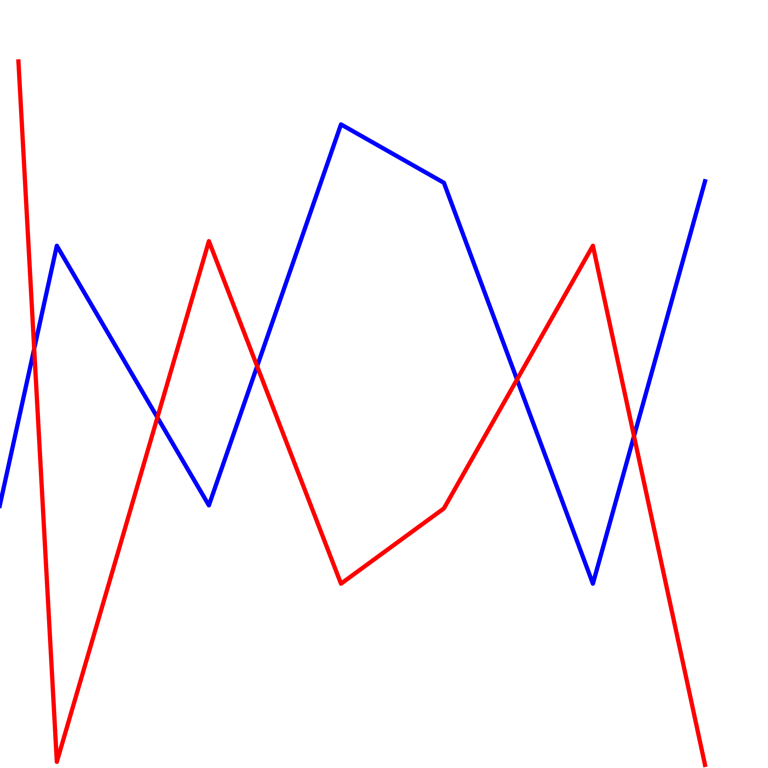}
  \caption{Number of intersections}
\end{subfigure}
\hfill
\begin{subfigure}[b]{0.23\textwidth}
  \includegraphics[width=\textwidth]{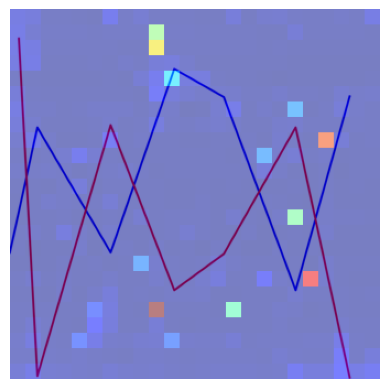}
  \caption{\# Intersections = \color{red}{1}}
\end{subfigure}
\hfill
\begin{subfigure}[b]{0.23\textwidth}
  \includegraphics[width=\textwidth]{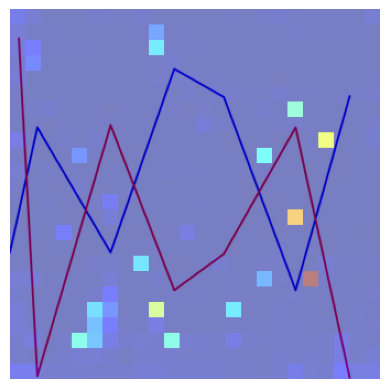}
  \caption{\# Intersections = \color{red}{3}}
\end{subfigure}
\hfill
\begin{subfigure}[b]{0.23\textwidth}
  \includegraphics[width=\textwidth]{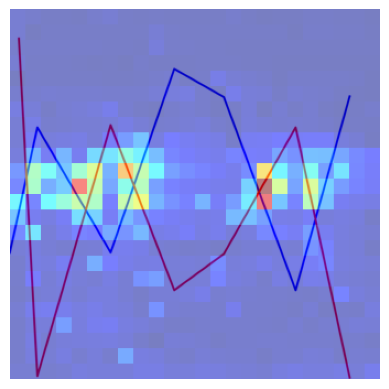}
  \caption{\# Intersections = \color{dgreen}{5}}
\end{subfigure}
\vspace{1em}
\setcounter{subfigure}{0}
\begin{subfigure}[b]{0.225\textwidth}
  \includegraphics[width=\textwidth]{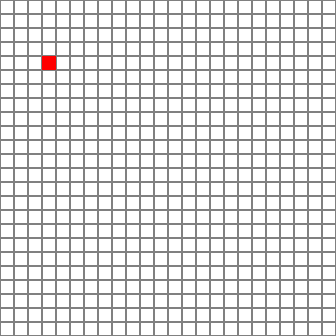}
  \caption{Where is the red square?}
\end{subfigure}
\hfill
\begin{subfigure}[b]{0.23\textwidth}
  \includegraphics[width=\textwidth]{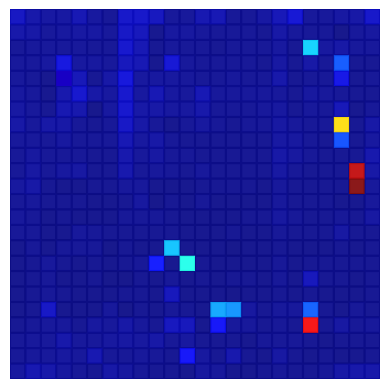}
  \caption{Model's response = \color{red}{(12, 12)}}
\end{subfigure}
\hfill
\begin{subfigure}[b]{0.23\textwidth}
  \includegraphics[width=\textwidth]{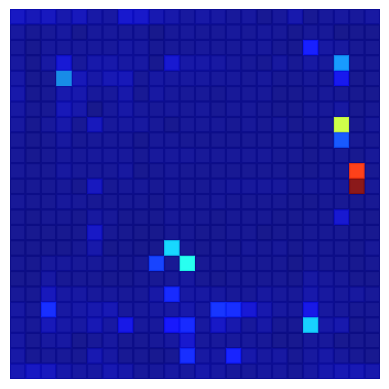}
  \caption{Model's response = \color{red}{(1, 4)}}
\end{subfigure}
\hfill
\begin{subfigure}[b]{0.23\textwidth}
  \includegraphics[width=\textwidth]{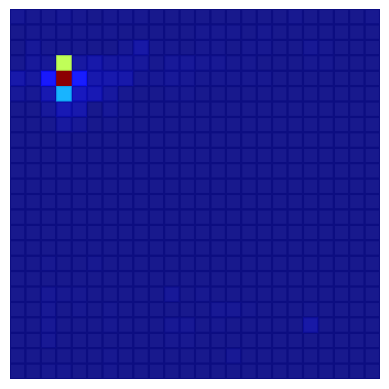}
  \caption{Model's response = \color{dgreen}{(3, 4)}}
\end{subfigure}
\vspace{1em}
\setcounter{subfigure}{0}
\begin{subfigure}[b]{0.23\textwidth}
  \includegraphics[width=\textwidth]{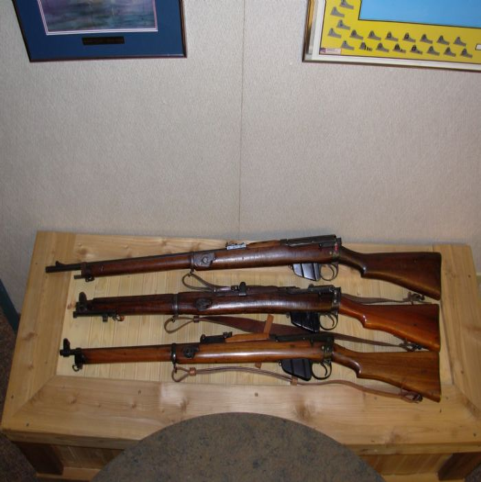}
  \caption{Where is the photo?}
\end{subfigure}
\hfill
\begin{subfigure}[b]{0.23\textwidth}
  \includegraphics[width=\textwidth]{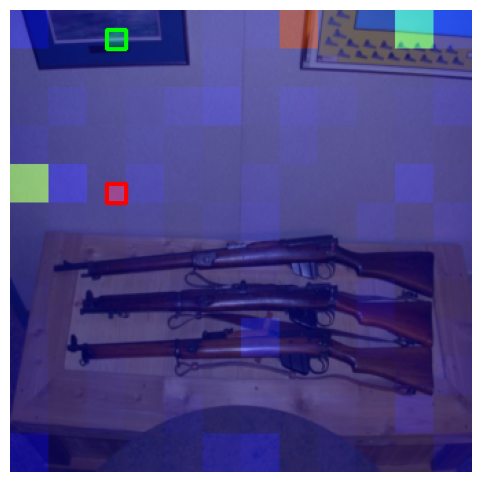}
  \caption{Predicted coords = \color{red}{(5, 9)}}
\end{subfigure}
\hfill
\begin{subfigure}[b]{0.23\textwidth}
  \includegraphics[width=\textwidth]{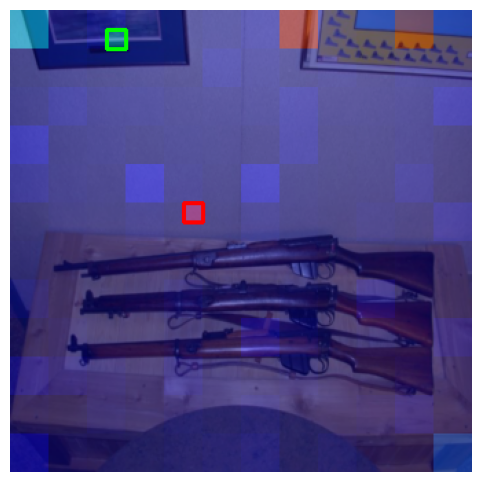}
  \caption{Predicted coords = \color{red}{(9, 10)}}
\end{subfigure}
\hfill
\begin{subfigure}[b]{0.23\textwidth}
  \includegraphics[width=\textwidth]{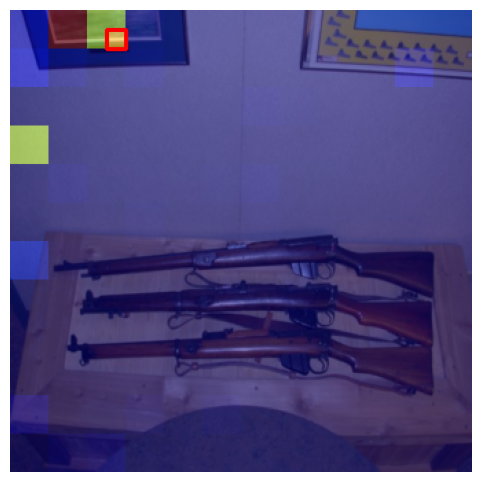}
  \caption{Predicted coords = \color{dgreen}{(5, 1)}}
\end{subfigure}
\vspace{1em}
\setcounter{subfigure}{0}
\begin{subfigure}[b]{0.21\textwidth}
  \includegraphics[width=\textwidth]{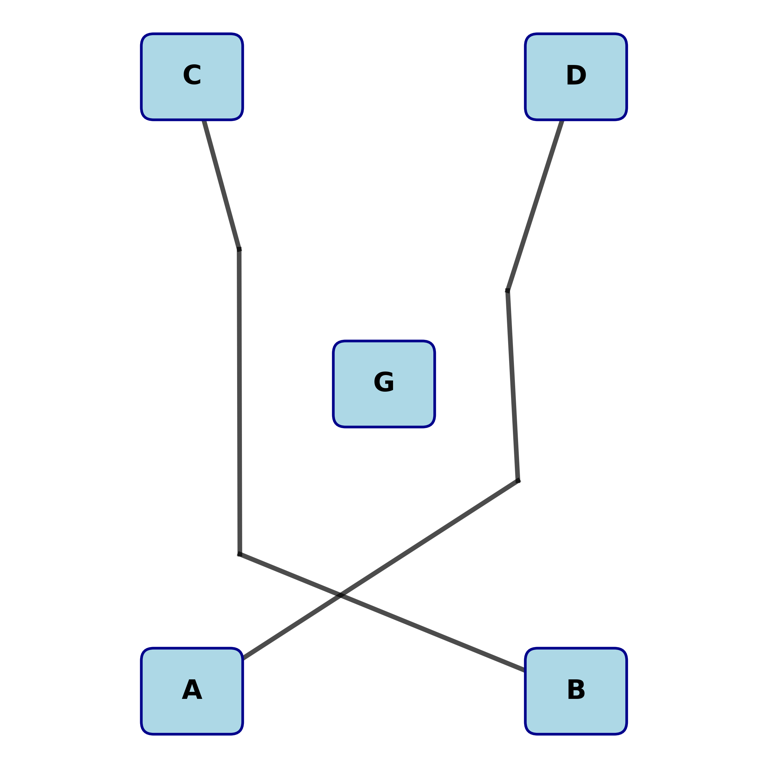}
  \caption{Is node A connected to D?}
\end{subfigure}
\hfill
\begin{subfigure}[b]{0.23\textwidth}
  \includegraphics[width=\textwidth]{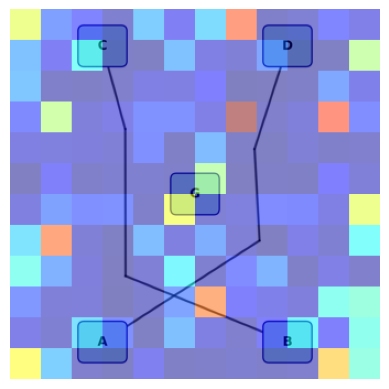}
  \caption{Model's response = \color{red}{"No"}}
\end{subfigure}
\hfill
\begin{subfigure}[b]{0.23\textwidth}
  \includegraphics[width=\textwidth]{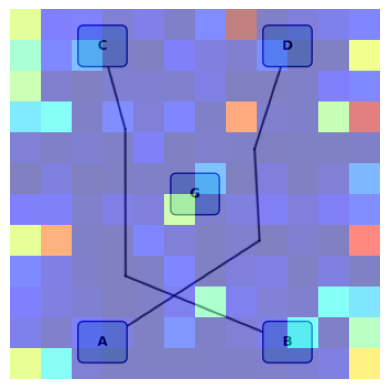}
  \caption{Model's response = \color{red}{"No"}}
\end{subfigure}
\hfill
\begin{subfigure}[b]{0.23\textwidth}
  \includegraphics[width=\textwidth]{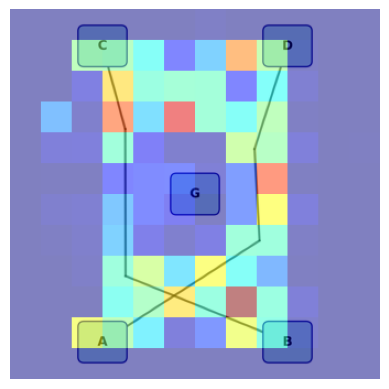}
  \caption{Model's response = \color{dgreen}{"Yes"}}
\end{subfigure}
\vspace{1em}
\caption{
Attention maps show the attention of the last answer token
to visual tokens. The results in the first two rows are from LLava-v1.5, and the last two rows show results from Qwen2.5-VL.
First column shows input images and tasks.
Second column shows the attention maps of out-of-the-box models. 
Third column shows the NTP-finetuned model.
Fourth column shows the attention maps of models finetuned with the proposed NTP+KLAL.
The red text indicates wrong answers and the green the correct ones. In the third row,
the green box indicates the ground truth patch, and the red box denotes the predicted patch.
Not only are the answers produced by NTP + KLAL correct, but the attention maps are also dramatically improved and become much more interpretable.
}
\label{fig:multi_task_attention}
\end{figure*}

\vspace{-2mm}
\section{Experimental Evaluation}
\vspace{-1mm}

\subsection{Datasets and Tasks} \label{sec:datasets}
We evaluated our method on \textcolor{blue}{five} datasets, each designed to test different aspects of spatial/geometric reasoning in VLMs. In the Line Intersection task, the model must count the number of intersections between lines, thereby testing its quantitative geometric understanding. In the Line Tracing task, it must determine whether two nodes are connected in a graph, assessing its capacity for tracing paths through visual structures. The two pointing tasks, one with synthetic images and one with real images, require the model to locate a target object and output its coordinates, evaluating the model's spatial localization and grounding capabilities. We also included a referring expression comprehension (REC) task using the well-established RefCOCO dataset \cite{yu2016modeling}, where the model must locate the object described in natural language by predicting the bounding box coordinates that contain it. For all datasets, we used an 80/20 train–test split, except for RefCOCO where we followed its standard splits.
\subsubsection{Geometric Datasets}
\textbf{Line Tracing:} We constructed a synthetic dataset of graph images with complex topologies to challenge the line tracing abilities of VLMs. 
As illustrated in the last row of Fig.~\ref{fig:multi_task_attention},
the graphs consist of a central node and 3 to 6 other labeled nodes positioned around it. 
Some nodes are connected with polygonal curves with both short and long-range connections. 
Each graph contains 2 or 3 disjoint edges, ensuring that every node is connected to only one other node and forms distinct, disconnected pairs. 
So, we ensured the graphs are not cluttered.
The dataset includes Yes/No questions asking whether two arbitrary nodes in a given graph are connected, resulting in 1,064 images and 5,360 question-answer pairs, with a balanced distribution of Yes and No answers.


\textbf{Line Intersection:} \cite{rahmanzadehgervi2024vision} introduced a geometric visual task to evaluate whether VLMs can count the number of intersections between two piecewise linear curves. The lines are colored in blue and red, 
with 0, 1, or 2 unique intersections. To increase difficulty, we extended the dataset by generating additional images where the number of intersections ranges from 3 to 5, e.g., see the first row in Fig.~\ref{fig:multi_task_attention}. So, the number of intersection points ranges from 0 to 5, with 200 images generated for each category, resulting in a total of 1,200 images labeled by the corresponding number of intersections.
The accompanying question is: “How many times do the blue and red lines touch each other?”
\subsubsection{Pointing Datasets}
The pointing task evaluates the ability of VLMs to localize a target object 
We consider two variants:
\textbf{Grid Patch:} Each image is divided into a $24{\times}24$ grid with gray overlay lines, and one target cell is highlighted in red. The prompt asks the model to output the grid coordinates of the red patch (see the second row of Fig.~\ref{fig:multi_task_attention}).
\textbf{PixMo-Points:} A subset of 1,500 real-world images from PixMo-Points~\cite{deitke2025molmo}, spanning 150 object categories. Each sample provides a short textual prompt and a human-annotated point marking the object center. The model must output the corresponding $(x,y)$ coordinates to the center of the referred object (see the third row of Fig.~\ref{fig:multi_task_attention}).

\vspace{-3mm}
\subsection{Results and Analysis}
\vspace{-2mm}

On geometric and pointing tasks (Line Intersection, Line Tracing, and object pointing), we evaluated the LLava-v1.5-7B and Qwen2.5-VL-7B-Instruct models under three configurations: (1) base model (out-of-the-box), (2) finetuned with the next token prediction (NTP), and (3) finetuned with NTP combined with the proposed KLAL (starting from base model check point). For the REC task on RefCOCO, we focused on Qwen2.5-VL-7B-Instruct, the stronger of the two models, and evaluated it under the same configurations.

We compared the two models against both open-source and commercial state-of-the-art VLMs. Among the commercial baselines, we included GPT‑4o \cite{hurst2024gpt} and Gemini-2.0 Flash \cite{comanici2025gemini}. 
For open-source baselines, we selected \textcolor{blue}{three} models. GLaMM-FS-7B~\cite{rasheed2024glamm}, specialized for grounding with region-level annotations, was evaluated only on the Line Intersection and Line Tracing tasks. Molmo-7B-D \cite{deitke2025molmo}, trained with coordinate-level supervision including the PixMo-Points (superset of our dataset), and InstructBLIP-Vicuna-7B \cite{dai2023instructblipgeneralpurposevisionlanguagemodels}, a general-purpose instruction-tuned model, were both evaluated across all geometric and pointing tasks.

Task-specific accuracy metrics were used for evaluation. For the Line Intersection task, a response was considered correct if it exactly matched the ground truth number, which ranged from 0 to 5. For the Line Tracing task, a response was deemed correct if it matched the ground truth answer, either Yes or No. For the two pointing tasks (Grid Patch and PixMo-Points), a prediction was considered correct if the predicted coordinates were within 3 units of Euclidean distance from the ground truth coordinates; otherwise, it was considered incorrect. 
Final accuracy was computed as the ratio of correct predictions to the total number of test samples for each dataset.

\begin{table}[!htbp]
  \centering
  \caption{Accuracy on Line Intersection and Line Tracing tasks}
  \label{tab:combined_line_accuracy}
  \begin{tabular}{lcc}
    \toprule
    \textbf{Method} &
      \textbf{Line Intersection} &
      \textbf{Line Tracing} \\
    \midrule
    \multicolumn{3}{l}{\textit{LLava-v1.5-7B}} \\
    \quad Base Model   & 27.91\% & 50.00\% \\
    \quad NTP          & 49.11\% & 46.76\% \\
    \quad NTP + KLAL     & 55.68\% & 53.52\% \\
    \midrule
    \multicolumn{3}{l}{\textit{Qwen2.5-VL-7B-Instr.}} \\
    \quad Base Model   & 47.62\% & 49.62\% \\
    \quad NTP          & 62.64\% & 53.82\% \\
    \quad NTP + KLAL     & \textbf{70.23\%} & \textbf{62.21\%} \\
    \midrule
    \multicolumn{3}{l}{\textit{SOTA}} \\
    \quad Molmo-7B-D         & 41.07\% & 49.51\% \\
    \quad GLaMM-FS-7B       & 27.50\% & 39.03\% \\
    \quad InstructBLIP      & 36.67\% & 46.23\% \\
    \quad GPT-4o            & 42.12\% & 55.34\% \\
    \quad Gemini-2.0 Flash  & 56.41\% & 59.25\% \\
    \bottomrule
  \end{tabular}
\end{table}

Table~\ref{tab:combined_line_accuracy} shows the out-of-the-box LLava-v1.5 and Qwen2.5-VL models performed above chance (16.7\%) on the Line Intersection task but struggled on the more challenging Line Tracing dataset, where they achieved near-random accuracy of 50\%. Finetuning with NTP led improved Qwen2.5-VL notably on both tasks. In contrast, LLava-v1.5 showed minimal gain on Line Tracing, likely due to its strong bias toward answering Yes. Adding KLAL to NTP resulted in significant gains: LLava-v1.5 improved by 6.6\% on Line Intersection and 6.8\% on Line Tracing, while Qwen2.5-VL improved by 7.6\% and 8.4\%, respectively. 
Although Gemini-2.0 Flash outperformed others, it still lagged behind Qwen2.5-VL with NTP + KLAL.
This is notable since Gemini-2.0 Flash and GPT-4o excel on more complex tasks.

\begin{table}[!htbp]
  \centering
  \caption{Accuracy on Grid Patch and PixMo-Points datasets. 
}
  \label{tab:accuracy_patch_pixmo}
  \begin{tabular}{lcc}
    \toprule
    \textbf{Method} &
      \textbf{Grid Patch} &
      \textbf{PixMo-Points} \\
    \midrule
    \multicolumn{3}{l}{\textit{LLava-v1.5-7B}} \\
    \quad Base Model          & 10.42\% &  5.84\% \\
    \quad NTP                 & 20.41\% &  9.49\% \\
    \quad NTP + KLAL            & 40.82\% & 16.52\% \\
    \midrule
    \multicolumn{3}{l}{\textit{Qwen2.5-VL-7B-Instr.}} \\
    \quad Base Model          &  6.12\% & 16.79\% \\
    \quad NTP                 & 28.57\% & 26.28\% \\
    \quad NTP + KLAL            & \textbf{44.90\%} & \textbf{35.77\%} \\
    \midrule
    \multicolumn{3}{l}{\textit{SOTA}} \\
    \quad Molmo‑7B‑D          & 18.37\% & 21.53\% \\
    \quad InstructBLIP    & 6.44\% & 8.31\% \\
    \quad GPT-4o              & 38.78\% & 19.70\% \\ 
    \quad Gemini-2.0 Flash    & 40.82\% &  18.98\% \\

    \bottomrule
  \end{tabular}
\end{table}

Table~\ref{tab:accuracy_patch_pixmo} shows results on the Grid Patch and PixMo-Points pointing tasks. In their base forms, both LLava-v1.5 and Qwen2.5-VL exhibited low accuracy, particularly on PixMo-Points (chance level is below 1\%, e.g., in Grid Patch a \(24 \times 24 = 576\) grid gives random accuracy \(1/576 \approx 0.17\%\)). The NTP + KLAL combination led to substantial improvements over NTP alone. LLava-v1.5 improved by 20.4\% on Grid Patch and 7.0\% on PixMo-Points, while Qwen2.5-VL gained 16.3\% and 9.5\%, respectively. Molmo-7B-D outperformed all commercial SOTA models on PixMo-Points, likely due to its finetuning in coordinate-level supervision on images from the same dataset. (Note that Molmo-7B-D was evaluated in its native pixel-level coordinates; we then linearly mapped its predictions into our grid coordinate system.) Across both datasets, Qwen2.5-VL consistently outperformed LLava-v1.5 and surpassed all baseline models when finetuned with NTP + KLAL.

To assess generalization, we conducted cross-dataset transfer experiments between PixMo-Points and Grid Patch to examine whether learning on one dataset can positively transfer to the other in Table~\ref{tab:transfer_results}. Interestingly, when transferring from Grid Patch to PixMo-Points, NTP performed worse than the base model, indicating that NTP alone cannot generalize to unseen datasets without large-scale training data. However, KLAL + NTP enabled Qwen2.5-VL to transfer effectively, achieving much higher accuracy and, in some cases, reaching performance comparable to task-specialized baselines such as Molmo.

\begin{table}[!htbp]
  \centering
  \caption{Transfer accuracy between Pixmo-Points (P) and Grid Patch (G) datasets. 
P $\rightarrow$ G: trained on P, evaluated on G. 
G $\rightarrow$ P: trained on G, evaluated on P.}
  \label{tab:transfer_results}
  \begin{tabular}{lcc}
    \toprule
    \textbf{Method} &
      \makecell{\textbf{P} $\rightarrow$ \textbf{G}} &
      \makecell{\textbf{G} $\rightarrow$ \textbf{P}} \\
    \midrule
    \multicolumn{3}{l}{\textit{Qwen2.5-VL-7B-Instr.}} \\
    \quad Base Model   & 6.12\%  & 16.79\% \\
    \quad NTP          & 11.16\% & 16.05\% \\
    \quad NTP + KLAL   & \textbf{19.29\%} & \textbf{21.98\%} \\
    \bottomrule
  \end{tabular}
\end{table}

To see whether our method scales to large, well-established visual grounding datasets, we evaluated KLAL on RefCOCO for the REC task. As shown in Table~\ref{tab:accuracy_refcoco}, while the base model and NTP already achieve strong accuracy, finetuning with KLAL + NTP yields further improvements across validation, testA, and testB splits. These results demonstrate that KLAL + NTP provides measurable gains even in high-performing settings and highlight its potential for broader application to real-world grounding benchmarks.

\begin{table}[!htbp]
  \centering
  \caption{Accuracy (IoU 0.5) of models finetuned on the RefCOCO training set and evaluated on RefCOCO validation and test splits.}
  \label{tab:accuracy_refcoco}
  \begin{tabular}{lccc}
    \toprule
    \textbf{Method} & \multicolumn{3}{c}{\textbf{RefCOCO}} \\
    \cmidrule(lr){2-4}
     & \textbf{val} & \textbf{testA} & \textbf{testB} \\
    \midrule
    \multicolumn{4}{l}{\textit{Qwen2.5-VL-7B-Instr.}} \\
    \quad Base Model          & 90.05\% & 93.75\% & 86.70\% \\
    \quad NTP                 & 90.65\% & 94.05\% & 86.90\% \\
    \quad NTP + KLAL          & \textbf{91.45\%} & \textbf{94.65\%} & \textbf{87.50\%} \\
    \bottomrule
  \end{tabular}
\end{table}

\vspace{-5mm}
\subsection{Visual Attention}

\begin{figure}[!htbp]
  \centering
    \includegraphics[width=0.7\columnwidth]{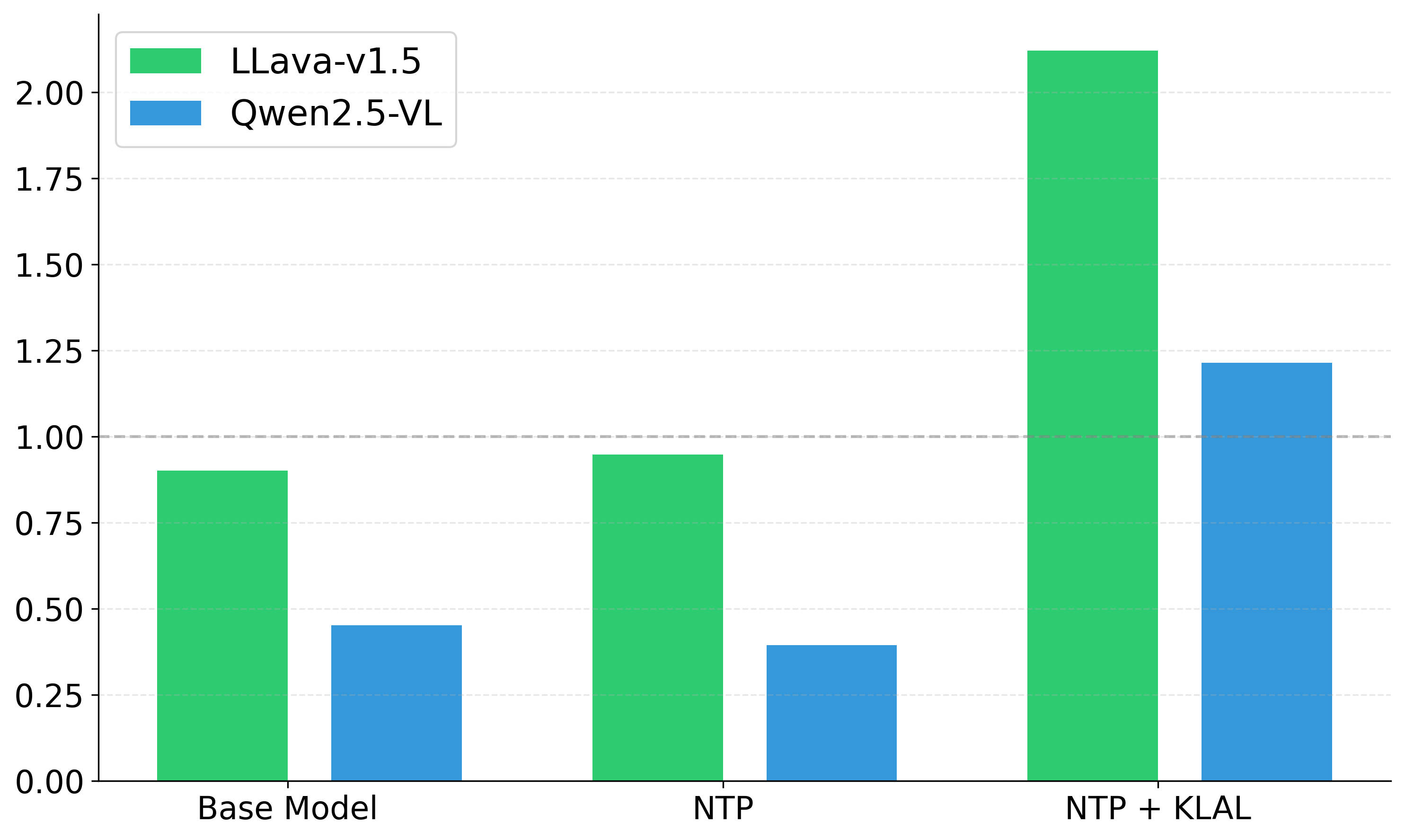}
    \vspace{-3mm}
  \caption{The bar plots illustrate the ratio of the average attention of target tokens w.r.t the answer token to the average attention of all visual tokens computed over all Line Intersection test images.
}
\label{fig:ratio_comparison}
\end{figure}

Across all datasets, KLAL with NTP improved accuracy, and enhanced attention maps as shown in Fig.~\ref{fig:multi_task_attention}.
We demonstrate this statistically in Fig.~\ref{fig:ratio_comparison} on the task of counting the number of intersection points, where the target visual patch tokens are those containing the intersection points.
The bar plots illustrate the ratio of the average attention of visual target tokens to the average attention of all visual tokens w.r.t the answer token.
The values below one for the base models and the NTP finetuned models indicate that the influence of the target tokens on the answer is smaller than the influence of other visual tokens.
\textbf{Only after finetuning with the proposed KLAL, we see that the average target token has higher influence on the answer than the average visual token.}

\begin{figure}[]
\centering
\setcounter{subfigure}{0}
\begin{subfigure}[b]{0.25\linewidth}
  \includegraphics[width=\linewidth]{figures/Line_Intersection/llava1.5_answer=1.png}
\end{subfigure}
\hfill
\begin{subfigure}[b]{0.25\linewidth}
  \includegraphics[width=\linewidth]{figures/Line_Intersection/llava_ntp_answer=3.png}
\end{subfigure}
\hfill
\begin{subfigure}[b]{0.25\linewidth}
  \includegraphics[width=\linewidth]{figures/Line_Intersection/llava_ntp+kl_answer=5.png}
\end{subfigure}
\vspace{1em}
\setcounter{subfigure}{0}
\begin{subfigure}[b]{0.25\linewidth}
  \includegraphics[width=\linewidth]{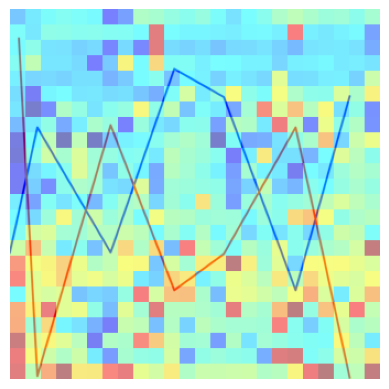}
  \caption{Out-of-the-box}
\end{subfigure}
\hfill
\begin{subfigure}[b]{0.25\linewidth}
  \includegraphics[width=\linewidth]
  {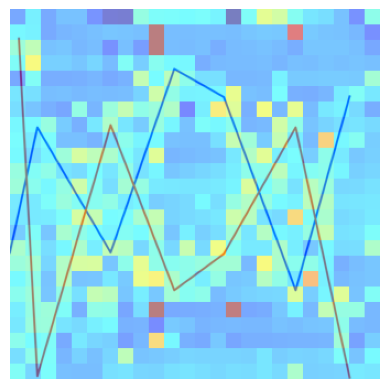}
  \caption{NTP}
\end{subfigure}
\hfill
\begin{subfigure}[b]{0.25\linewidth}
  \includegraphics[width=\linewidth]{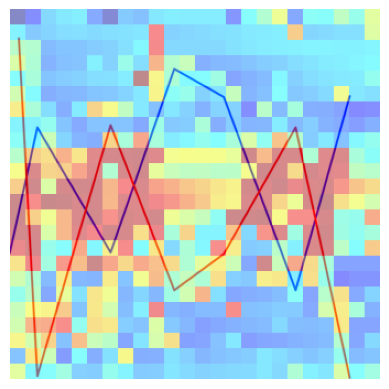}
  \caption{NTP + KLAL}
\end{subfigure}
\vspace{-5mm}
\caption{
The first row shows attention maps from the last output token of LLava-v1.5, while the second row presents a visualization derived from the magnitudes of visual token embeddings. Both types of visualizations reveal similar patterns, as enhanced attention through NTP + KLAL corresponds to increased focus of high-norm tokens on the target regions.
}
\label{fig:vectornormmaps}
\end{figure}

Interestingly, our attention-focused loss function also positively influences the deep embeddings of visual tokens.  
In the second row of Fig.~\ref{fig:vectornormmaps}, we replaced the attention of visual tokens to the answer token with the norm of the embeddings of visual tokens obtained from the last attention layer.
As we can see, the proposed NTP + KLAL significantly increased the concentration of high-norm tokens at the line intersection points. This correlates with the attention to the answer token shown in the first row.
Since high-norm tokens are more likely to be attended to, they are more likely to influence the answer.
The higher concentration of high-norm visual tokens at the target regions is also confirmed by the results in
Table~\ref{tab:ratio_norm_values}. 
It demonstrates that using NTP alone has little impact on embedding magnitude, whereas adding KLAL increases the average norm by 6\% for Qwen2.5-VL and by 19\% for LLava-v1.5. This confirms that KLAL not only directs model attention but also strengthens the internal feature representations of the relevant visual tokens.

\begin{table}[!htbp]
  \centering
  \footnotesize
  \setlength{\tabcolsep}{4pt}      
  \renewcommand{\arraystretch}{0.9} 
  \begin{tabular}{lcc}
    \toprule
    \textbf{Model}    & \textbf{NTP} & \textbf{KLAL+NTP} \\
    \midrule
    LLava-v1.5        & 1.05              & \textbf{1.19}                    \\
     Qwen2.5-VL        & 0.96              & \textbf{1.06}                    \\
    \bottomrule
  \end{tabular}
  \vspace{-3mm}
\caption{Ratio of average embedding norm of target visual tokens after finetuning to the base model on the line intersection test set.}
  \label{tab:ratio_norm_values}
\end{table}

\vspace{-3mm}

To summarize our experimental results clearly demonstrate our main hypothesis that direct supervision of the attention of language tokens to corresponding visual tokens in the LLM module of VLMs leads to improved performance on visual tasks.
We demonstrated this on four simple but fundamental visual tasks as well as on the real-world RefCOCO benchmark.
The details of the finetuning process are mentioned in Section 4 of the supplementary material.

\vspace{-3mm}
\section{Related Work}
\vspace{-2mm}

Large-scale VLMs such as CLIP~\cite{radford2021learning} and Flamingo~\cite{alayrac2022flamingo} align image and text via contrastive learning or frozen encoders with autoregressive LMs. 
We focus on instruction-tuned, autoregressive VLMs~\cite{li2023blip, liu2023visual, bai2023qwenvlversatilevisionlanguagemodel, zhu2023minigpt, lu2024deepseek} that generate text or special tokens as output.
These VLMs have achieved impressive results on tasks ranging from image captioning~\cite{li2023blip, wang2021simvlm}, image–text retrieval~\cite{radford2021learning,yu2022coca}, to visual question answering~\cite{alayrac2022flamingo}.  
The successes largely stem from training on massive image–text corpora, which improve overall alignment between modalities.  Yet scale alone does not guarantee faithful grounding: VLMs regularly hallucinate plausible but incorrect details when no image is provided~\cite{payattentionimageECCV2024} and can ignore up to half of the visual tokens without degrading performance~\cite{chen2024imageworth12tokens}, a symptom often traced to so-called “attention sinks” that absorb excessive weight despite low semantic relevance~\cite{kang2025attentionsink}. Moreover, existing benchmarks may not sufficiently test vision-centric abilities of VLMs ~\cite{tong2024cambrian, tong2024eyes}. 
This shows that while larger and more diverse image–text datasets may boost benchmark scores, they do not ensure that generated text attends to the correct image regions.


To address these gaps, one approach integrates VLMs with visual grounding, mapping expressions to image regions by finetuning on grounding datasets and adding localization modules. Examples include LISA~\cite{lai2024lisa} with a `[SEG]` token and mask decoder, F-LMM~\cite{wu2025f} with lightweight mask refinement, and GLAMM~\cite{rasheed2024glamm} with an added grounding encoder and pixel decoder.


Another approach modifies attention only at inference, e.g., boosting visual token weights~\cite{payattentionimageECCV2024}, redistributing mass from sink tokens~\cite{kang2025attentionsink}, or pruning redundant tokens~\cite{chen2024imageworth12tokens}. These reduce hallucination but leave representations unchanged and underperform finetuned models ~\cite{kang2025your}.

A third line of work supervises cross-modal attention during training. Inspired by attention–rationale alignment in NLP~\cite{wiegreffe2019attention} and Grad-CAM in vision~\cite{selvaraju2017grad}, recent VQA and grounding works add auxiliary losses that guide tokens toward annotated or automatically generated maps~\cite{parelli2023interpretable, VisualGroundingCVPR2024}. Other approaches include attention regularization~\cite{liu2022answer} and attention priors or multi-grained grounding for more relevant visual focus~\cite{le2023guiding, huang2019multi}.
These typically rely on object detectors, saliency maps, or external grounding models.
Closely related to our goal, FastRM~\cite{stan2025fastrmefficientautomaticexplainability} and FiVL~\cite{aflalo2025fivlframeworkimprovedvisionlanguage} analyze and encourage vision–language alignment by examining or leveraging attention patterns, with an emphasis on explainability and evaluation. Our KL Attention Loss (KLAL) follows this line of work; however, it differs by automatically deriving task-specific GT attention maps from underlying task properties or provided annotations and aligning visual-to-answer attention via layer-wise KL divergence.

Unlike approaches such as Pix2Seq~\cite{pix2seqICLR2022,kosmos2-ICLR2023}, which train LLMs to output bounding box coordinates (transferring visual knowledge into text), we directly link language tokens to relevant visual tokens. Our focus is not global embedding alignment as in CLIP~\cite{radford2021learning}, UNITER~\cite{chen2020uniter}, or ViLBERT~\cite{lu2019vilbert}, but token-level grounding of answer tokens. Experiments show this is difficult with the standard NTP loss, as LLM attention layers seem to be unable to properly connect language and visual tokens when guided only by NTP.



While VLMs excel at general image understanding, they often struggle with geometric and relational reasoning on abstract structures such as line perception and connectivity \cite{rahmanzadehgervi2024vision, li2025graph}. To address this, we introduce a synthetic Line Tracing dataset (Fig.~\ref{fig:multi_task_attention}) testing node connectivity via visual paths in complex graphs. Unlike chart-QA datasets \cite{kahou2017figureqa, masry2022chartqa} or broader geometric benchmarks \cite{kazemi2023geomverse}, it targets visual path perception, providing a benchmark for assessing ability to follow polygonal curves and infer topological connectivity.

\vspace{-4mm}
\section{Discussion and Conclusions}
\vspace{-2mm}

The proposed approach is inspired by
direct supervision of visual attention in early childhood learning. 
As is well-established in psychology \cite{tomasello1986joint,wood1976role},
direct supervision during tasks like object identification in pictures is not just helpful, but it is crucial for child development. 
It transforms passive exposure into active learning.
In particular, linking visual attention and word learning is essential \cite{yu2012embodied}:
\emph{
``moments in which a single object was visually dominant. If parents named the object during these moments of bottom-up selectivity, later forced-choice tests showed that infants learned the name, but did not when naming occurred during a less visually selective moment."}
Therefore, we propose a novel loss function that directly supervises the attention of language tokens to corresponding vision tokens,
and demonstrate experimentally its benefits.

Our approach is model-agnostic, integrates into existing VLMs without architectural changes, and requires no task-specific heads (e.g., for localization or segmentation). While we focused on visual tasks fundamental to spatial understanding, the proposed approach can be extended to other complex visual tasks where no annotations are provided.

\vspace{-4mm}
\section{Acknowledgment}
\vspace{-2mm}
This work was supported by NSF grant IIS-2331768.

{
    \small
    \bibliographystyle{ieeenat_fullname}
    \bibliography{references}
}

\end{document}


\pagestyle{empty}

\maketitle
\thispagestyle{empty}

\section{Datasets}
Below, we provide more information on dataset the creation steps and ground truth attention maps.

\subsection{Line Tracing}
\begin{figure}[t!]
  \centering
  \begin{subfigure}[b]{0.48\columnwidth}
    \includegraphics[width=\textwidth]{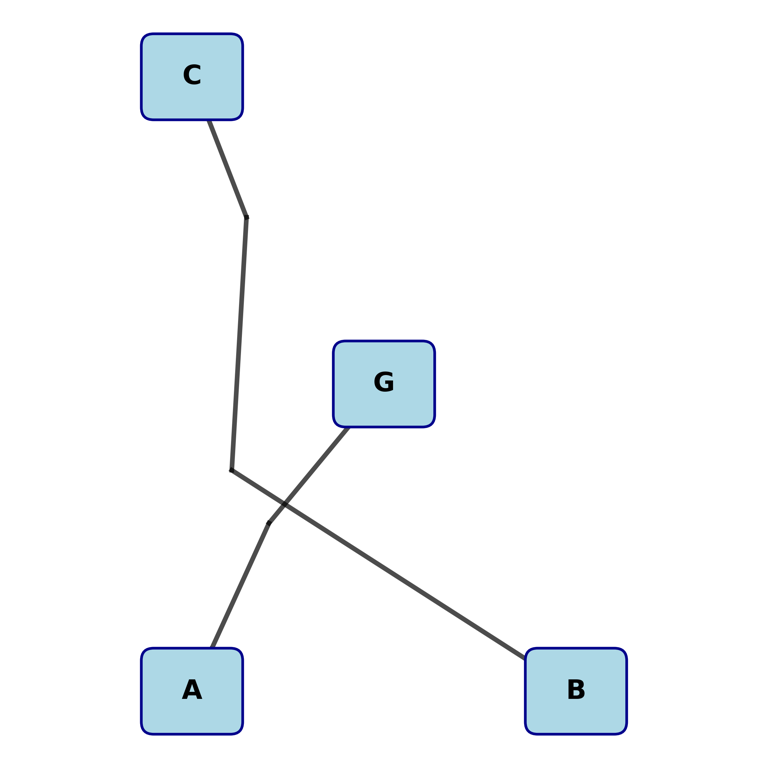}
    \caption{Is node A connected to node G?}
  \end{subfigure}\hfill
  \begin{subfigure}[b]{0.48\columnwidth}
    \includegraphics[width=\textwidth]{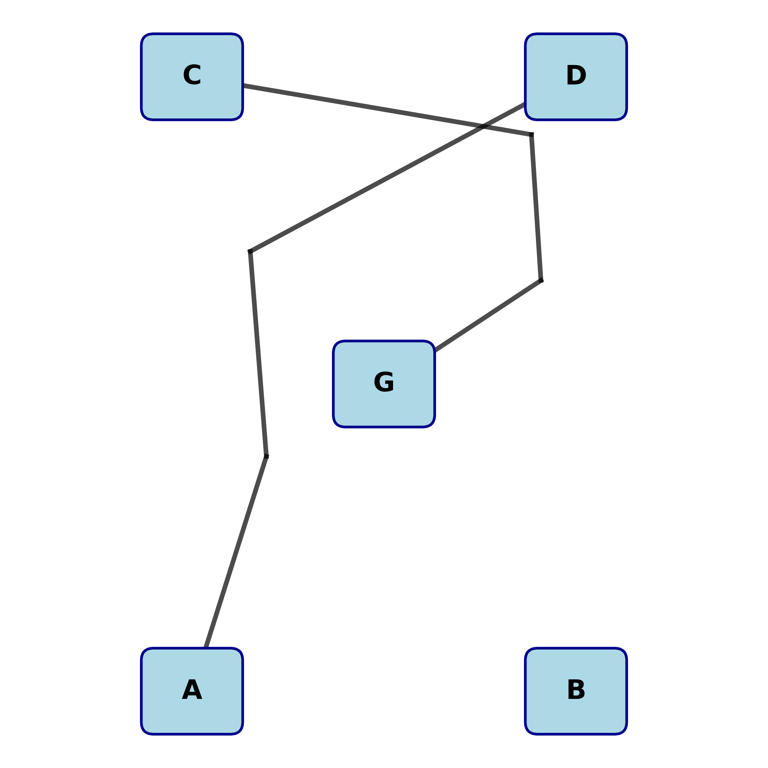}
    \caption{Is node A connected to node C?}
  \end{subfigure}

  \vspace{0.5em}

  \begin{subfigure}[b]{0.48\columnwidth}
    \includegraphics[width=\textwidth]{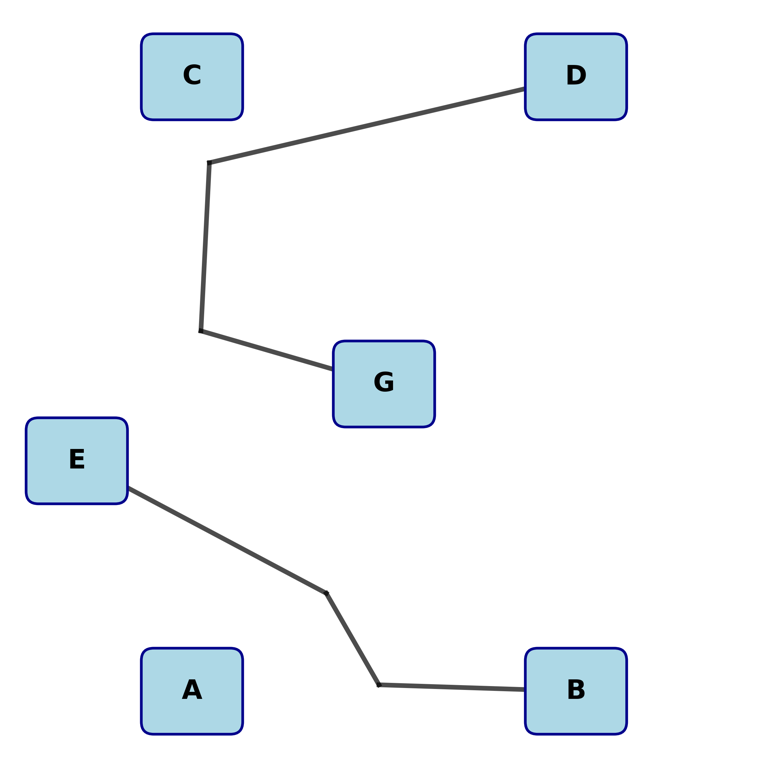}
    \caption{Is node B connected to node E?}
  \end{subfigure}\hfill
  \begin{subfigure}[b]{0.48\columnwidth}
    \includegraphics[width=\textwidth]{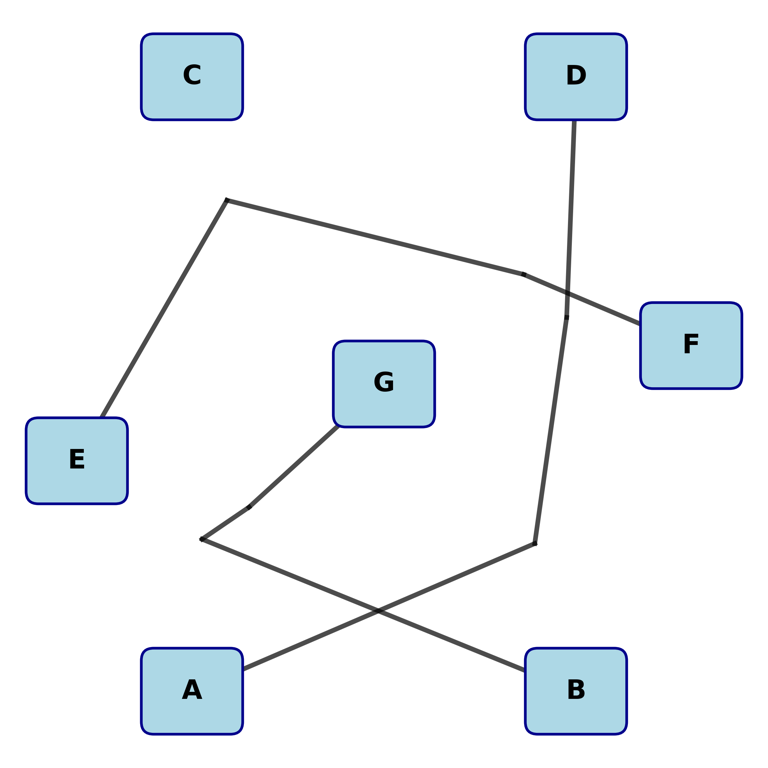}
    \caption{Is node E connected to node D?}
  \end{subfigure}

  \caption{Sample input images from the synthetically generated Line Tracing dataset, each paired with a Yes/No connectivity question.}
  \label{fig:line_tracing}
\end{figure}

As shown in Fig.\ref{fig:line_tracing}, Graph nodes were drawn as labeled boxes (labels A–F) placed at fixed grid positions, with edges selected by sampling node pairs whose Euclidean distance exceeded a minimum threshold and then rendered as colored polygonal curves (line width set for clear visibility). Each vertex of these curves received a small random jitter to avoid overly regular layouts. To increase diversity without changing connectivity, every unique graph topology was rendered under simple augmentations (horizontal flip, vertical flip, and 90° rotation), producing multiple visual variants per structure.

\subsection{Line Intersection} 
\begin{figure}[t!]
  \centering
  \begin{subfigure}[b]{0.16\textwidth}
    \includegraphics[width=\linewidth]{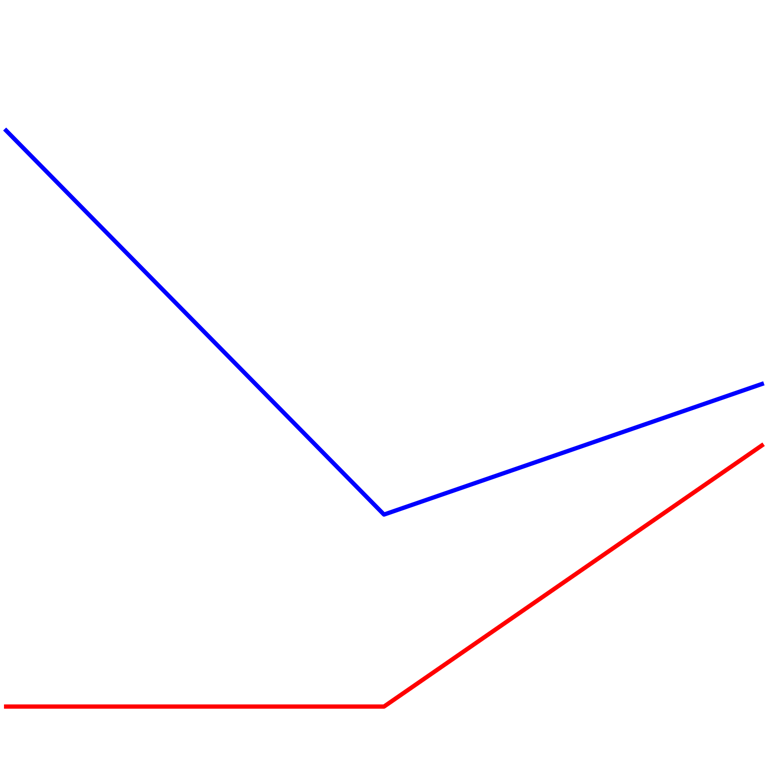}
    \caption{Label: 0}
    \label{fig:int-0}
  \end{subfigure}
  \begin{subfigure}[b]{0.16\textwidth}
    \includegraphics[width=\linewidth]{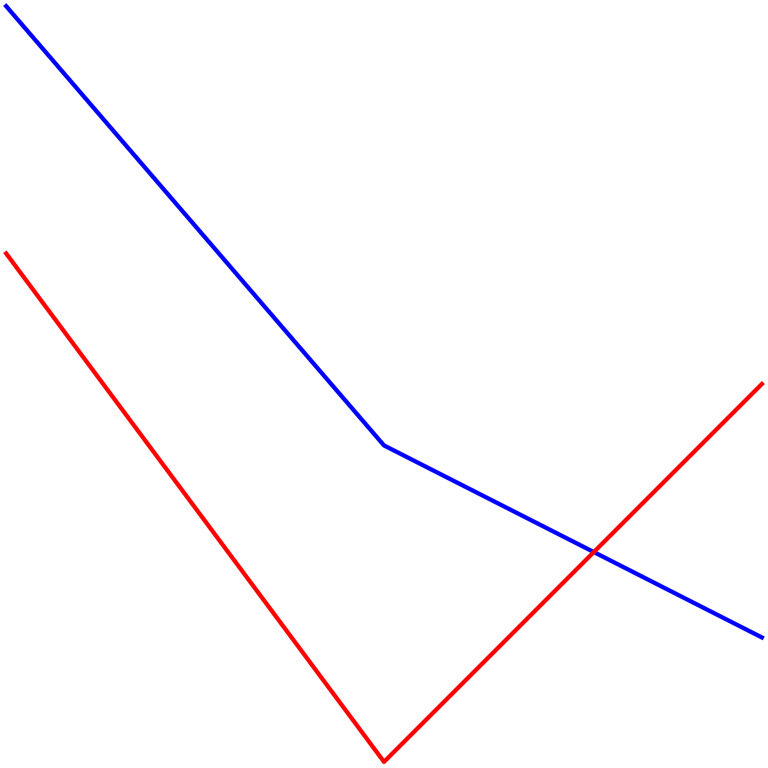}
    \caption{Label: 1}
    \label{fig:int-1}
  \end{subfigure}
  \begin{subfigure}[b]{0.16\textwidth}
    \includegraphics[width=\linewidth]{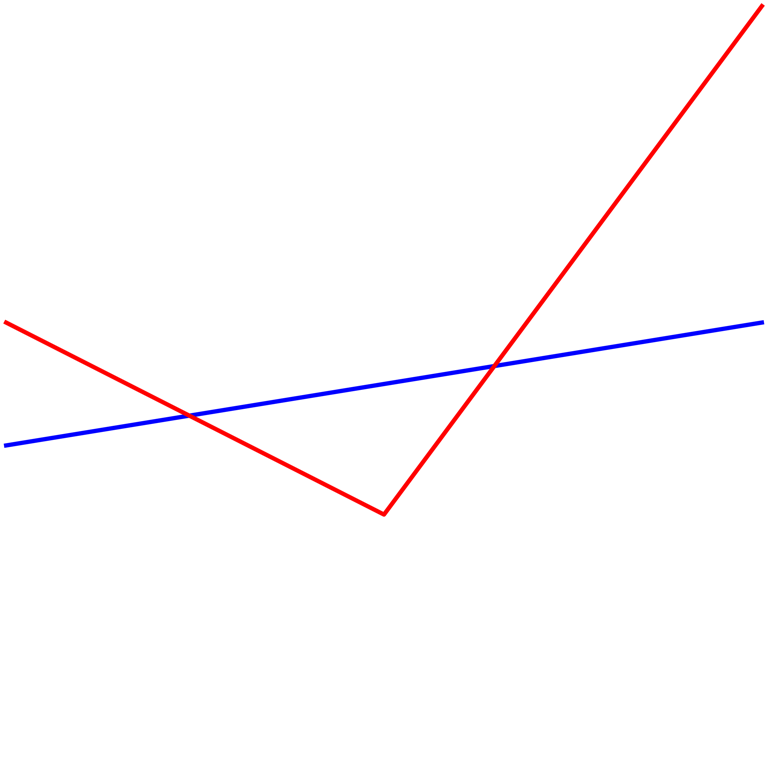}
    \caption{Label: 2}
    \label{fig:int-2}
  \end{subfigure}
  \begin{subfigure}[b]{0.16\textwidth}
    \includegraphics[width=\linewidth]{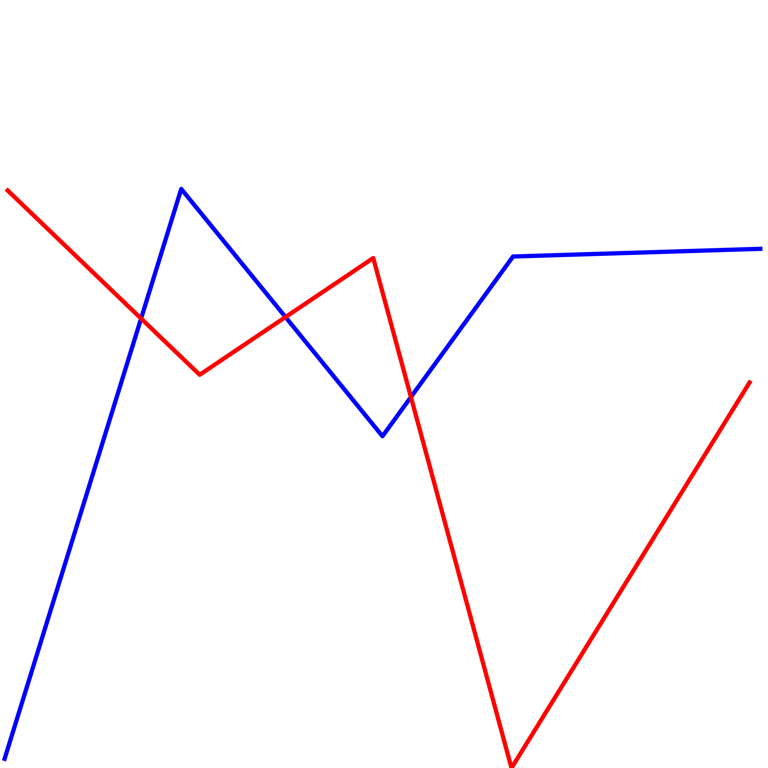}
    \caption{Label: 3}
    \label{fig:int-3}
  \end{subfigure}
  \begin{subfigure}[b]{0.16\textwidth}
    \includegraphics[width=\linewidth]{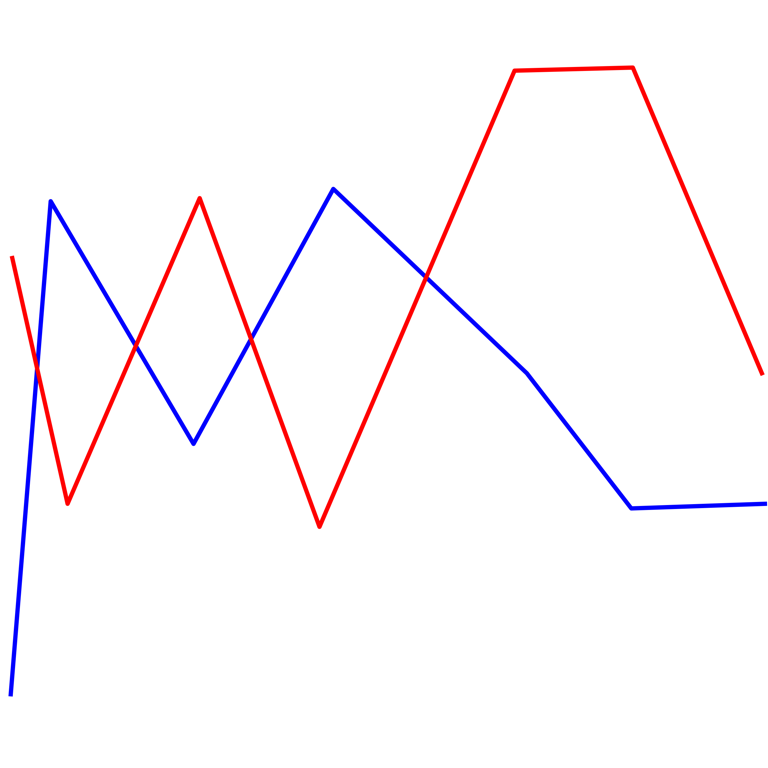}
    \caption{Label: 4}
    \label{fig:int-4}
  \end{subfigure}
  \begin{subfigure}[b]{0.16\textwidth}
    \includegraphics[width=\linewidth]{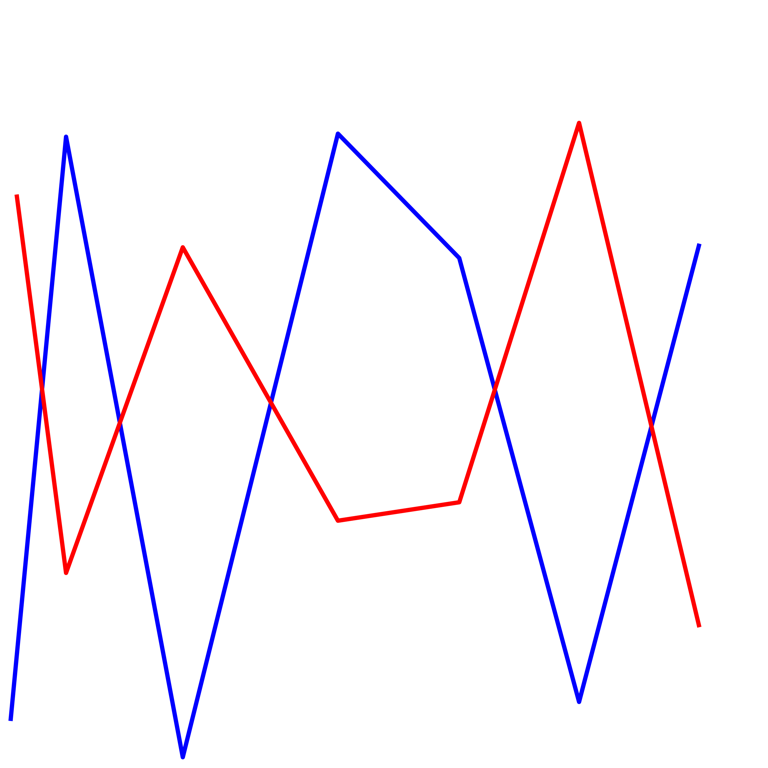}
    \caption{Label: 5}
    \label{fig:int-5}
  \end{subfigure}
  \caption{Sample images from the synthetically generated Line Intersection dataset with corresponding intersection counts between the red and blue lines (possible values: 0–5).}
  \label{fig:line_intersection}
\end{figure}

Each image was created by first laying out a 12 by 12 grid of potential vertex positions, then constructing two piecewise linear curves whose complexity depends on the target intersection count. For zero to two intersections, each curve had three points at the left, middle, and right positions, while for three to five intersections additional randomly jittered control points were added. The vertical positions of these points were sampled from predefined bands (low, mid, high) to steer how often the curves cross, and we computed exact line segment intersections, discarding and retrying any pair that did not match the desired count. Once a valid pair of red and blue curves was found, it was rendered as an image and paired with its correct intersection label. Figure~\ref{fig:line_intersection} shows one representative image for each intersection count (0–5) to illustrate the variety in our dataset.

\subsection{Grid Patch}
Grid Patch is a synthetic dataset in which each image is rendered on a white background and divided into a 24×24 grid with gray overlay lines. The target patch, occupying one grid cell, is highlighted in red.  
The prompt asks for the grid coordinates of the red patch in the form $(x, y)$, after defining the coordinate system: (0, 0) as the top-left and (23, 23) as the bottom-right cell.  
Specifically, $x$ and $y$ correspond to the column and row indices within the grid.  
The dataset contains 576 synthetic images, each paired with a question, with individual grid cells matching image patches and measuring 14×14 pixels.

\subsection{PixMo-Points Subset}
The PixMo-Points dataset is a large-scale resource for visual grounding based on real-world images. Human annotators identified objects, provided textual descriptions, and pointed to the center of each object instance within the images. For our experiments, we used a subset of 1,500 images covering 150 randomly selected object categories. The task involves identifying the center of the referred object and predicting its coordinates $(x, y)$, following the same format as used in the Grid Patch dataset.

\subsection{Target Patch Selection}
In this section, we explain how we select target patches for each dataset. Once we have target patches, we can generate ground truth attention maps as explained in the methodology section.\\
\textbf{Line Tracing.} We sampled points uniformly along each edge segment and assigned each sample to the nearest grid patch by comparing its coordinates to patch‐center positions. We also ensured that the patches containing the edge’s endpoints were always included.\\
\textbf{Line Intersection.} We treated each polygonal curve as a set of line segments, detected their crossings using orientation tests and exact intersection formulas, computed the intersection coordinates, and removed any repeated points so that each intersection is only counted once. The patches covering these unique points are the targets.\\ 
\textbf{Grid Patch and PixMo‐Points.} We took the provided ground truth coordinates and directly mapped them to patch indices in the grid, using each coordinate pair as a target visual token.\\
\textbf{RefCOCO.}To find the centerline within the bounding box, we first check whether the box is taller or wider. The smaller of the two dimensions (width or height) is used as a reference to keep the axis well within the bounding box. If the box is taller, a vertical line is drawn, centered horizontally using an x-offset from the left edge. If the box is wider, a horizontal line is drawn, centered vertically using a y-offset from the top edge. In both cases, the line is shortened slightly on each end to avoid touching the edges directly. The target patches are defined as those traversed by this centerline.

\section{Patch and PixMo-Points Performance}
Table~\ref{tab:combined_dataset_performance} reports results of all models on the Grid Patch and PixMo-Points datasets under base, NTP, and NTP+KLAL configurations. Here, in addition to accuracy (where predictions within three grid units of the ground truth count are correct), we also include the median distance errors.

\begin{table*}[!htbp]
  \centering
  \caption{Performance comparison of models on Patch and Pixmo datasets}
  \label{tab:combined_dataset_performance}
  \begin{tabular}{lcccc}
    \toprule
    \textbf{Method} &
      \multicolumn{2}{c}{\textbf{Patch Dataset}} &
      \multicolumn{2}{c}{\textbf{Pixmo Dataset}} \\
    \cmidrule(lr){2-3} \cmidrule(lr){4-5}
     & \textbf{Median Distance} & \textbf{Accuracy} & \textbf{Median Distance} & \textbf{Accuracy} \\
    \midrule
    \multicolumn{5}{l}{\textit{LLaVA-v1.5-7B}} \\
    \quad Base Model          & 8.06  & 10.42\% & 13.60 &  5.84\% \\
    \quad NTP                 & 5.39  & 20.41\% &  9.43 &  9.49\% \\
    \quad NTP + KLAL            & \textbf{3.61}  & 40.82\% &  7.14 & 16.52\% \\
    \midrule
    \multicolumn{5}{l}{\textit{Qwen2.5-VL-7B-Instruct}} \\
    \quad Base Model          & 11.05 &  6.12\% &  7.07 & 16.79\% \\
    \quad NTP                 & 6.08  & 28.57\% &  6.08 & 26.28\% \\
    \quad NTP + KLAL                & 4.00  & \textbf{44.90}\% &  \textbf{5.00} & \textbf{35.77}\% \\
    \midrule
    \multicolumn{5}{l}{\textit{SOTA}} \\
    \quad Molmo‑7B‑D          & 6.71  & 18.37\% &  7.13 & 21.53\% \\
    \quad GPT-4o              & 4.26  & 38.78\% &  7.07 & 19.70\% \\ 
    \quad Gemini-2.0 Flash    & 4.16  & 40.82\% &  7.21 &  18.98\% \\
    \bottomrule
  \end{tabular}
\end{table*}

\section{Attention Analysis}
This section examines how visual attention is distributed across the LLM’s layers and how it transforms when projected from the vision encoder into the model’s language embedding space. 
\subsection{Layer‐wise Visual Attention}
\begin{figure}[htbp]
  \centering
  \begin{subfigure}[b]{0.48\textwidth}
    \includegraphics[width=\textwidth]{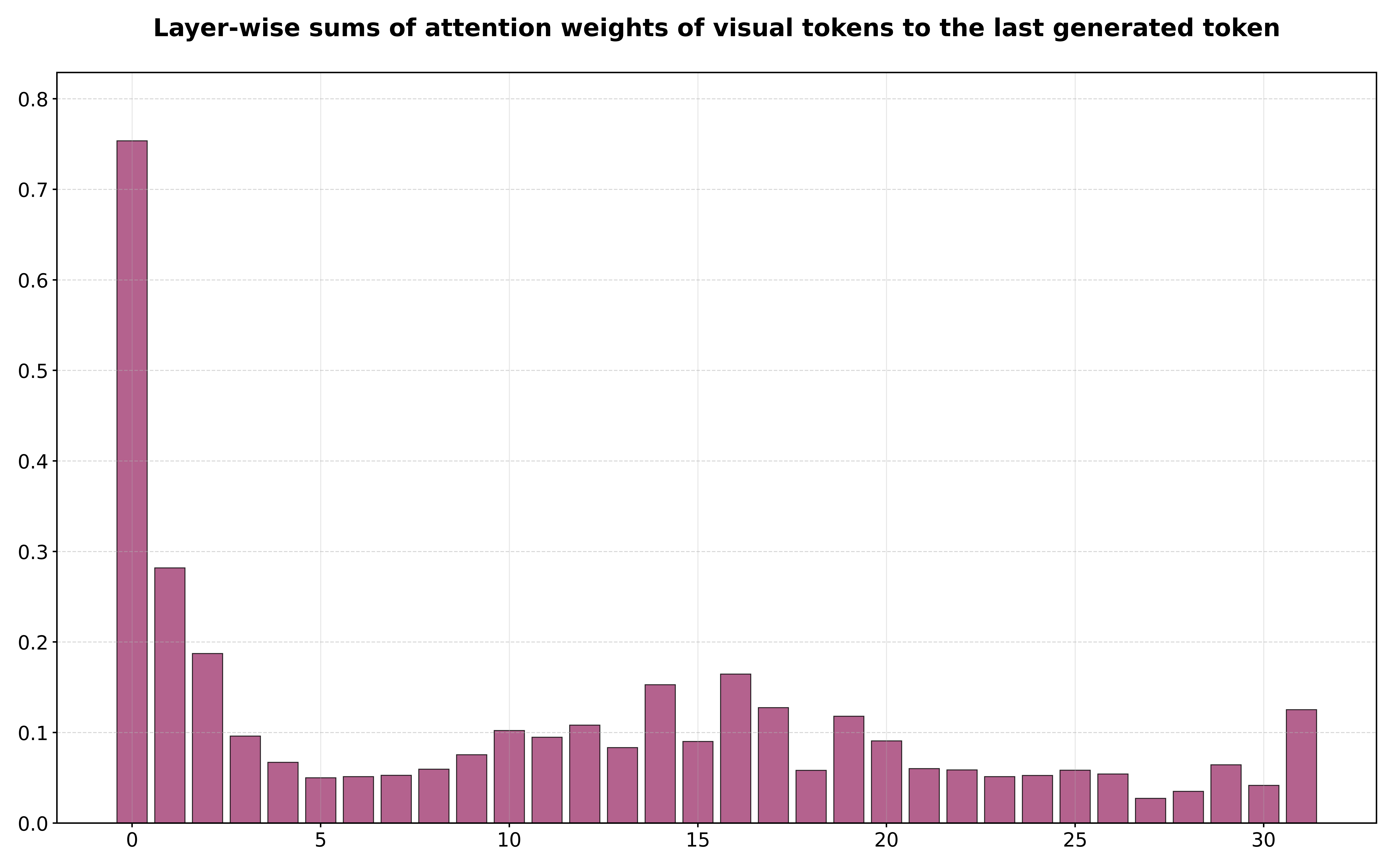}
    \caption{LLava-v1.5-7B Base Model}
    \label{fig:attention-llava}
  \end{subfigure}
  \quad
  \begin{subfigure}[b]{0.48\textwidth}
    \includegraphics[width=\textwidth]{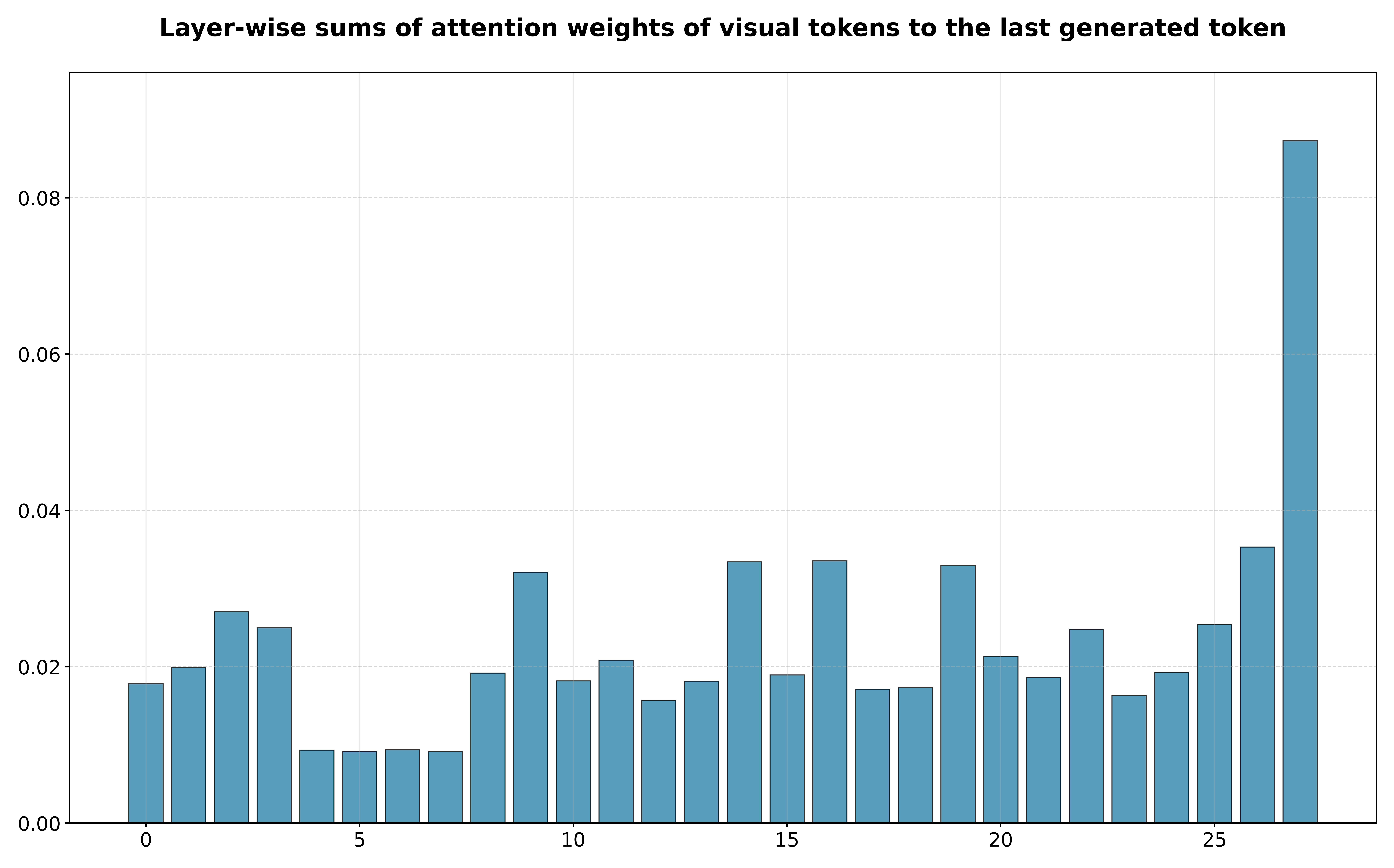}
    \caption{Qwen2.5-VL-7B-Instruct Base Model}
    \label{fig:attention-qwen}
  \end{subfigure}
  \caption{Layer-wise visual attention analysis (sum of attention weights to visual tokens) for LLava-v1.5 and Qwen-2.5VL.}
  \label{fig:layerwise-attention}
\end{figure}

Fig.~\ref{fig:layerwise-attention} compares the per‐layer sum of attention weights to visual tokens for the LLava‐v1.5 and Qwen2.5‐VL base models. Across layers, the attention of the last answer token to visual tokens comprises about 10.93\%(SD ± 12.66\%) of the total attention to all tokens in LLava-v1.5, which is substantially higher than Qwen-2.5VL’s average of 2.33\% (SD ± 1.44\%).

\subsection{Vision Encoder vs. LLM Decoder Attention} 
\begin{figure}[H]
    \centering
    \begin{subfigure}[b]{0.21\textwidth}
        \includegraphics[width=\textwidth]{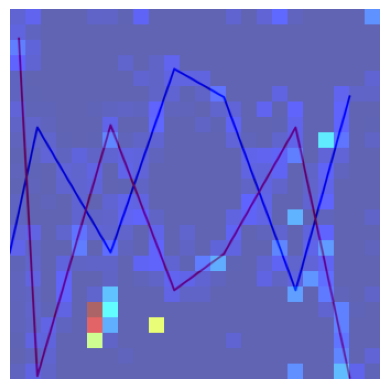}
        \caption{Vision Encoder}
    \end{subfigure}
    \begin{subfigure}[b]{0.21\textwidth}
        \includegraphics[width=\textwidth]{figures/Line_Intersection/llava1.5_answer=1.png}
        \caption{LLM Decoder}
    \end{subfigure}
    \caption{(a) The Vision Encoder tracks the lines. (b) LLM loses that focus and ignores the intersection points.}
    \label{fig:vision-vs-llm}
\end{figure}

Fig.~\ref{fig:vision-vs-llm} shows that the Vision Encoder maintains partial focus on the lines, whereas the LLM loses that focus in LLava-v1.5.

\section{Impact of Layer, Head, and Weighting Choices}
We conduct ablation studies to assess the design choices in our KLAL loss. 
Table~\ref{tab:klal_ablation_layers_heads} compares strategies for aggregating attention signals across layers and heads. 
Applying the loss to each layer, with attention matrices averaged across heads, achieves the highest accuracy, whereas restricting supervision to specific layers or heads leads to a performance drop. 
For selecting the top 5 heads, we considered two measures: (i) the sum of attention weights assigned to visual tokens, and (ii) the sum of attention weights assigned to the top 20 patches in the ground truth attention maps. 
We then ranked heads according to both measures and selected the five that overlapped across the two rankings.

Table~\ref{tab:lambda_ablation} examines the influence of the weighting hyperparameter $\lambda$ in the combined loss 
$\mathcal{L}_{\text{total}} = \mathcal{L}_{\text{NTP}} + \lambda \mathcal{L}_{\text{KLAL}}$. 
Performance peaks at $\lambda = 1.0$, indicating that balanced contributions from both terms are most effective.

\begin{table}[!htbp]
  \centering
  \small
  \caption{Ablation of design choices for attention supervision loss on Line Intersection for \textit{Qwen2.5-VL-7B-Instr.} with \textbf{NTP + KLAL}. Unless otherwise noted, results are averaged across heads within each layer.}
  \label{tab:klal_ablation_layers_heads}
  \begin{tabular}{lc}
    \toprule
    \textbf{Aggregation strategy} & \textbf{Accuracy (\%)} \\
    \midrule
    All layers              & \textbf{70.23\%} \\
    Top 4 layers            & 68.73\% \\
    Middle layers (13--16)  & 63.00\% \\
    Last layer only         & 65.43\% \\
    Each head independently (all layers) & 67.26\% \\
    Top 5 heads only (all layers)        & 64.84\% \\
    \bottomrule
  \end{tabular}
\end{table}

\begin{table}[!htbp]
  \centering
  \small
  \caption{Effect of varying the hyperparameter $\lambda$ in the combined loss 
  ($\mathcal{L}_{\text{total}} = \mathcal{L}_{\text{NTP}} + \lambda \mathcal{L}_{\text{KLAL}}$) 
  on Line Intersection for \textit{Qwen2.5-VL-7B-Instr.}}
  \label{tab:lambda_ablation}
  \begin{tabular}{lc}
    \toprule
    \textbf{$\lambda$} & \textbf{Accuracy (\%)} \\
    \midrule
    0.25 & 67.03\% \\
    0.50 & 68.98\% \\
    0.75 & 69.96\% \\
    1.00 & \textbf{70.23\%} \\
    1.25 & 68.86\% \\
    1.50 & 67.77\% \\
    \bottomrule
  \end{tabular}
\end{table}

\section{Implementation Configuration}

We finetuned the LLava-v1.5-7B and Qwen2.5-VL-7B-Instruct VLMs using DeepSpeed on two NVIDIA A6000 GPUs (40 GB VRAM each). The vision encoder remained frozen except for a trainable projection layer, while Low-Rank Adaptation (LoRA) was applied to the language component: LLava-v1.5 employed a LoRA rank of 16 (after rank 8 proved insufficient), and Qwen2.5-VL used a LoRA rank of 8. Training was conducted with a per-GPU batch size of 32 (effective global batch size = 64). Optimization followed a cosine decay learning rate schedule (initial LR = $5\times10^{-5}$) with a linear warm-up over the first 3\% of total training steps. 
For the REC task on RefCOCO, we finetuned Qwen2.5-VL-7B-Instruct for 400 steps under the same optimization setup described above.

For the commercial VLM baselines (GPT-4o and Gemini-2.0 Flash), we queried the public APIs using their default system prompts. We performed greedy decoding, did not apply top-$k$ or top-$p$ sampling, and left the temperature parameter at its default setting, ensuring that outputs were deterministic and directly comparable to those from our finetuned open-source models.